\documentclass[journal]{IEEEtran}

\usepackage{xcolor,soul,framed} 

\colorlet{shadecolor}{yellow}
\usepackage[pdftex]{graphicx}
\graphicspath{{../pdf/}{../jpeg/}}
\DeclareGraphicsExtensions{.pdf,.jpeg,.png}

\usepackage[cmex10]{amsmath}
\usepackage{array}
\usepackage{mdwmath}
\usepackage{mdwtab}
\usepackage{eqparbox}
\usepackage{amsfonts,amssymb}
\usepackage{url}
\usepackage{cite} 
\usepackage{multirow}
\usepackage{multicol}
\usepackage{color}
\definecolor{blu}{RGB}{0,0,255}

\begin{document}
\bstctlcite{IEEEexample:BSTcontrol}
    \title{Open-Vocabulary Remote Sensing Image  Semantic Segmentation}
  \author{Qinglong~Cao,~\IEEEmembership{Graduate Student Member,~IEEE,}
      Yuntian~Chen,~\IEEEmembership{Member,~IEEE,}\\
      Chao~Ma,~\IEEEmembership{Member,~IEEE,}
      and~Xiaokang~Yang,~\IEEEmembership{Fellow,~IEEE}

  \thanks{This work was supported in part by the National Science Foundation of China under Grants 62106116, and in part by the National Science Foundation of China under Grant
62071388, and in part by China Meteorological Administration Climate Change Special Program (CMA-CCSP) under Grant QBZ202316.}
  \thanks{Q. Cao is with the MoE Key Lab of Artificial Intelligence, AI Institute, Shanghai Jiao Tong University, Shanghai 200240, China, and also with the Ningbo Institute of Digital Twin, Eastern Institute of Technology, Ningbo 315200, China (e-mail: caoql2022@sjtu.edu.cn).}
  \thanks{Y. Chen is with Ningbo Institute of Digital Twin, Eastern Institute of Technology, Ningbo 315200, China, and also with the School of Electronic Information and Electrical Engineering, Shanghai Jiao Tong University 200240, China (e-mail: ychen@eitech.edu.cn).}%
  \thanks{C. Ma and X. Yang are with the MoE Key Lab of Artificial Intelligence, AI Institute, Shanghai Jiao Tong University, Shanghai 200240, China.}
  
  \thanks{\textit{Corresponding author: Yuntian Chen}}
  }  

\markboth{IEEE TRANSACTIONS ON GEOSCIENCE AND REMOTE SENSING
}{Cao \MakeLowercase{\textit{et al.}}: Open-Vocabulary Remote Sensing Image Semantic Segmentation}

\maketitle

\begin{abstract}

Open-vocabulary image semantic segmentation (OVS) seeks to segment images into semantic regions across an open set of categories. Existing OVS methods commonly depend on foundational vision-language models and utilize similarity computation to tackle OVS tasks. However, these approaches are predominantly tailored to natural images and struggle with the unique characteristics of remote sensing images, such as rapidly changing orientations and significant scale variations. These challenges complicate OVS tasks in earth vision, requiring specialized approaches. To tackle this dilemma, we propose the first  OVS framework specifically designed for remote sensing imagery, drawing inspiration from the distinct remote sensing traits. Particularly, to address the varying orientations, we introduce a rotation-aggregative similarity computation module that generates orientation-adaptive similarity maps as initial semantic maps. These maps are subsequently refined at both spatial and categorical levels to produce more accurate semantic maps. Additionally, to manage significant scale changes, we integrate multi-scale image features into the upsampling process, resulting in the final scale-aware semantic masks. To advance OVS in earth vision and encourage reproducible research, we establish the first open-sourced OVS benchmark for remote sensing imagery, including four public remote sensing datasets. Extensive experiments on this benchmark demonstrate our proposed method achieves state-of-the-art performance. All codes and datasets are available at https://github.com/caoql98/OVRS.

\end{abstract}


%
\IEEEpeerreviewmaketitle


\begin{figure*}
  \begin{center}
  \includegraphics[width=1.0\linewidth]{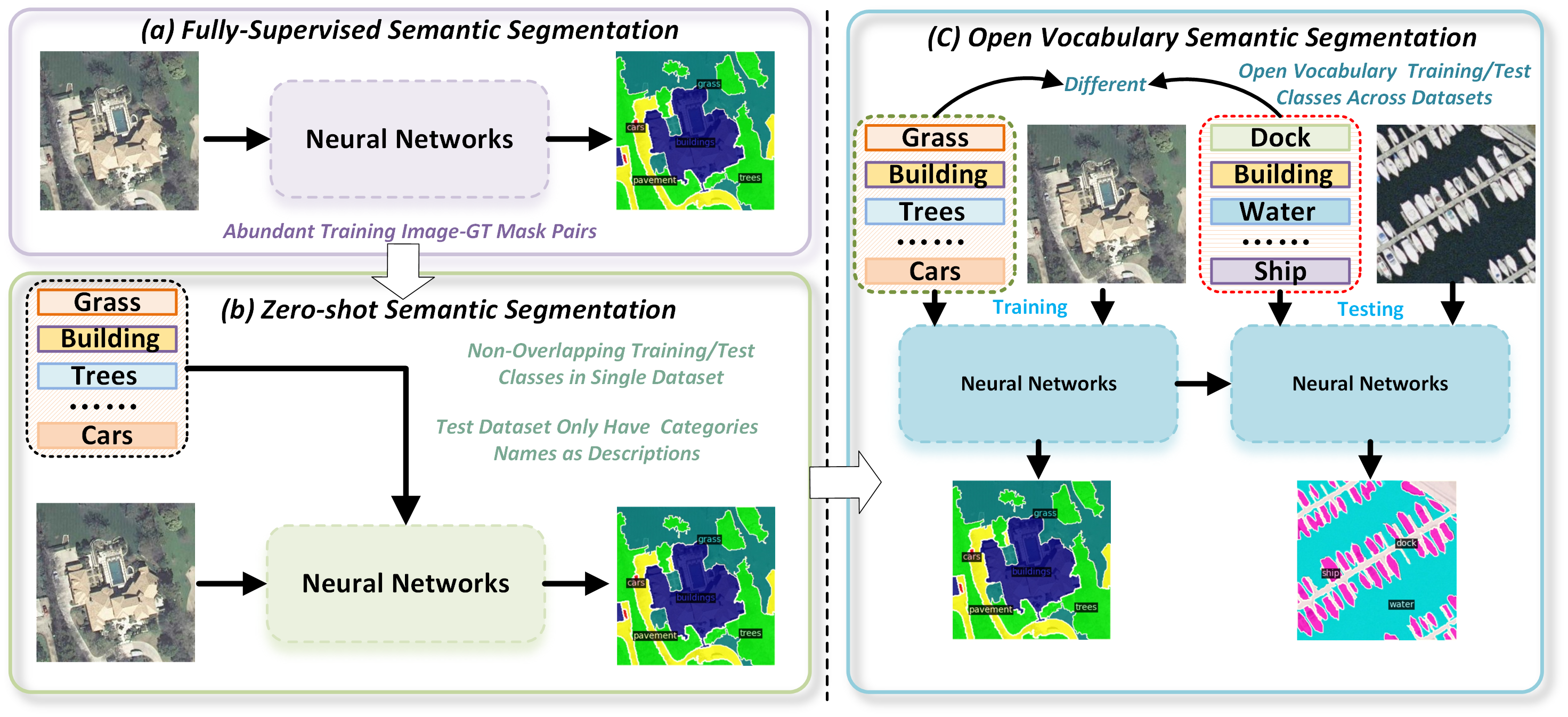}\\
  \caption{The differences between diverse semantic segmentation tasks for remote sensing images. (a) Fully supervised segmentation required abundant annotated data. (b) Zero-shot segmentation requires only class names for test datasets in a closed set. (c)  Open-vocabulary segmentation could achieve open-set segmentation across datasets with only category names.}
  \label{fig1}
  \end{center}
\end{figure*}


\section{Introduction}


\IEEEPARstart{S}{emantic} segmentation of remote sensing imagery is a crucial task aimed at the precise pixel-level classification of various elements within an image, which plays a pivotal role in numerous applications~\cite{chen2017deeplab,zhao2017pyramid,diakogiannis2020resunet,cao2023break}. The advent of deep learning has revolutionized this field, enabling fully supervised segmentation techniques~\cite{wang2024samrs,lv2023deep,hong2023cross} to achieve remarkable accuracy and efficiency. These methods have been effectively employed in diverse domains, including detailed land-use and land-cover mapping~\cite{forster1985examination,jat2008monitoring,ru2021land}, comprehensive traffic monitoring systems~\cite{zhao2022satsot,macioszek2021extracting,jian2019combining}, and the accurate extraction of infrastructure such as buildings and roads~\cite{qiao2023weakly,ding2021non,jian2019combining}. Despite their success, these fully supervised approaches are highly dependent on the availability of large-scale, well-annotated datasets~\cite{deng2009imagenet,everingham2010pascal,lin2014microsoft}. The extensive manual labeling required for such datasets is not only time-consuming and costly but also introduces a significant limitation: the models trained on these datasets often overfit to the specific categories seen during training, which results in poor generalization to unseen categories.

As illustrated in Figure~\ref{fig1} (b), to tackle the challenge of segmenting unseen categories, many researchers have introduced zero-shot segmentation algorithms~\cite{jiao2023learning,ding2022decoupling,bucher2019zero}. These algorithms rely solely on class names to infer the semantic regions for categories not present in the training datasets yet in a closed category set. Despite these advancements, such methods remain confined to individual datasets. Consequently, when deployed in real-world environments where models encounter categories beyond those included in their closed category set, their performance often deteriorates significantly. This performance decline highlights a critical limitation in the generalization capability of these models, ultimately restricting their usefulness in dynamic or heterogeneous environments where the range of potential categories is extensive and unpredictable.

To overcome this limitation and extend the applicability of segmentation models to more realistic scenarios, open-vocabulary image semantic segmentation (OVS) has been proposed, as depicted in Figure~\ref{fig1} (c).  OVS aims to segment remote sensing images by considering an open set of categories across different datasets. Existing OVS methods mainly utilize the generalized vision-language models like CLIP~\cite{radford2021learning} as the foundational models, and utilize the similarities between the features of images and category names to judge the semantic regions. For instance, Liang~\textit{et al.}~\cite{liang2023open} propose a two-stage OVS models, which firstly generate the mask proposals and subsequently segment the semantic regions based on the similarities of mask-category pairs.  Similarly, Freeseg~\cite{qin2023freeseg} adopts the two-stage framework and further introduces prompt learning for the text encoder, which helps the network perform the OVS tasks in the one-in-all pattern. Following the prompt learning manner, SegPrompt successfully leverages the category information as prompts to improve the model’s class-agnostic segmentation ability for the OVS tasks.  Though these OVS have achieved some success, they all focus on natural images. which could not handle the unique traits of remote sensing imagery, such as the rapidly changing orientations and the significant scale variations. These problems would obviously hamper the OVS performance for remote sensing imagery and seriously impede the development of open-vocabulary earth-vision tasks. 

To address the challenges inherent in open-vocabulary semantic segmentation of remote sensing imagery, we introduce a novel algorithm designed to be the first OVS framework for remote sensing imagery. One of the primary obstacles in remote sensing is the rapid change in object orientations, which poses a significant challenge for traditional segmentation methods. We posit that an effective solution requires a mechanism capable of adaptively capturing and aggregating semantic information across various orientations, thereby minimizing the impact of these variations. To this end, we propose a rotation-aggregative similarity computation module that synthesizes orientation-adaptive similarity maps to serve as the foundation for initial semantic maps. Our method begins by rotating the input remote sensing images across multiple orientations. These rotated images are then processed through the vision branch of a pre-trained CLIP model to extract orientation-varying image features. Concurrently, the language branch of CLIP generates class-specific features for the target categories. By computing the similarity between the orientation-specific image features and the class features, the network produces a set of orientation-varying similarity maps. These maps are subsequently aggregated across all orientations, yielding rotation-adaptive similarity maps that form the initial semantic maps. These initial maps are then subjected to further refinement at both spatial and categorical levels, enhancing their accuracy and robustness.

In addition to addressing orientation variability, our algorithm tackles the significant scale variations typical of remote sensing imagery by leveraging features from multiple levels of the feature extraction network. Each level of the network captures image features at different scales, which are crucial for accurate segmentation. During the upsampling process, these multiscale features are progressively integrated into the semantic maps, enriching them with comprehensive scale-related information and resulting in the generation of scale-aware semantic maps. Specifically, at each stage of upsampling, the semantic maps are averaged in both channel and spatial dimensions to produce semantic activation vectors. These vectors are then used to selectively activate the image features, which are subsequently combined with the semantic maps to form progressively refined, scale-aware semantic representations. This multiscale integration strategy ensures that the final semantic maps are not only orientation-adaptive but also scale-aware, thus significantly improving the segmentation performance across diverse and dynamic remote sensing scenarios.

By addressing the unique challenges of remote sensing imagery, we introduce a novel approach to open-vocabulary segmentation in earth vision. The key contributions of this paper are summarized as follows:

\begin{itemize}
	\item We propose the first open-vocabulary semantic segmentation framework tailored specifically for remote sensing imagery, eestablishing a new benchmark for remote sensing OVS and advancing the research in earth vision.
	
	\item To address the challenges of rapidly changing orientations and significant scale variations, we introduce a rotation-aggregative similarity computation module and progressively generate scale-aware semantic maps through the integration of multiscale features, resulting in more accurate and robust segmentation.
	
	\item Extensive experiments on our newly proposed remote sensing OVS benchmark demonstrate that our method significantly outperforms existing state-of-the-art natural image-based OVS approaches, effectively handling the distinct characteristics of remote sensing imagery.
	
\end{itemize}

\section{Related Work}

In this section, we first review related work on remote sensing image semantic segmentation. Then, the advanced research with regard to open-vocabulary semantic segmentation is extensively introduced. 

\subsection {Remote Sensing Image Semantic Segmentation}
Remote sensing image semantic segmentation focuses on the precise delineation of semantic regions within remote sensing imagery~\cite{yuan2021review, cao2023few}. The advent of deep learning has significantly advanced this field, leading to the development of numerous sophisticated algorithms. For example, Li~\textit{et al.} developed efficient attention modules to capture contextual dependencies, enhancing segmentation accuracy. Building on this, SSAtNet~\cite{zhao2021semantic} introduced a pyramid attention module that leverages multiscale features for adaptive refinement of segmentation features. Prioritizing computational efficiency, LANet~\cite{ding2020lanet} introduced patch attention and local embeddings to achieve effective segmentation through a patch-focused approach, while ResU-Net~\cite{li2021multistage} incorporated a linear attention mechanism to improve computational efficiency over traditional dot-product attention. Focusing on spatial information, HRCNet~\cite{xu2020hrcnet} developed a high-resolution context extraction network to better capture global contextual information, thereby enhancing segmentation performance. The transformer architecture~\cite{vaswani2017attention}, known for its superior image understanding capabilities across various computer vision tasks, has also been adapted for remote sensing image segmentation. For instance, Li~\textit{et al}~\cite{xu2021efficient} adopt the efficient transformer to perform the semantic segmentation task. Based on this concept, UNetFormer~\cite{wang2022unetformer} constructs the Transformer-based decoder and further proposes an UNet-like Transformer (UNetFormer) for real-time remote sensing segmentation.  Inspired by the powerful global modeling capabilities of the swin transformer, He~\textit{et al}~\cite{he2022swin} adopt swin transformer to mine pixel-level correlation to enhance the feature representation ability. To boost the global information extracted by the transformer network with local information from the convolutional neural network (CNN), zhang~\cite{zhang2022transformer} proposed a transformer and CNN  hybrid segmentation network, where the transformer is adopted for long-range spatial dependencies modeling, and CNN is utilized to maintain the local details. Similarly, STransFuse~\cite{gao2021stransfuse} jointly leveraged the Swin Transformer and CNN to extract coarse-grained and fine-grained feature representations and perform a staged semantic segmentation at diverse semantic scales.

\subsection{Open-Vocabulary Semantic Segmentation}
Open-vocabulary semantic segmentation seeks to accurately delineate semantic regions across an unrestricted set of categories. Current methods primarily utilize foundational vision-language models like CLIP~\cite{radford2021learning}, leveraging the similarities between textual category names and image features to identify corresponding semantic regions. For instance, Liang~\textit{et al.}\cite{liang2023open} fine-tune CLIP using paired masked image regions and text descriptions, enabling efficient classification of these masked regions. CLIPseg\cite{luddecke2022image} follows a prompt learning approach, directly using text descriptions as prompts to segment query images. Building on this adapter-based approach, SAN~\cite{xu2023side} introduces a side adapter network that generates mask proposals and attention biases, guiding the deeper layers of CLIP for proposal-wise classification. Similarly, SegCLIP~\cite{luo2023segclip} introduces the concept of super-pixels into OVS, aggregating image patches around learnable centers to form semantic regions based on pre-trained CLIP features. Extending this patch aggregation approach, Chen~\textit{et al.}~\cite{chen2023exploring} perform OVS by summarizing localized regions of the target image and distilling visual concepts using CLIP models. Focusing on the semantic alignment between visual content and unbounded text, SCAN~\cite{liu2024open} incorporates a generalized semantic prior and a contextual shift strategy to enhance segmentation performance. In contrast, SED~\cite{xie2024sed} addresses the often-overlooked local spatial information by utilizing a CNN-based CLIP to construct an efficient OVS network. More directly, CAT-Seg~\cite{cho2024cat} explores the multi-modal nature established between image and text embeddings, performing segmentation through cost volume computation. However, all these methods are constructed based on natural images and fail to address the unique characteristics of remote sensing imagery. To tackle this challenge and draw inspiration from these distinct traits, we propose the first open-vocabulary semantic segmentation framework specifically designed for remote sensing images.


\begin{figure*}[ht!]
  \begin{center}
  \includegraphics[width=1.0\linewidth]{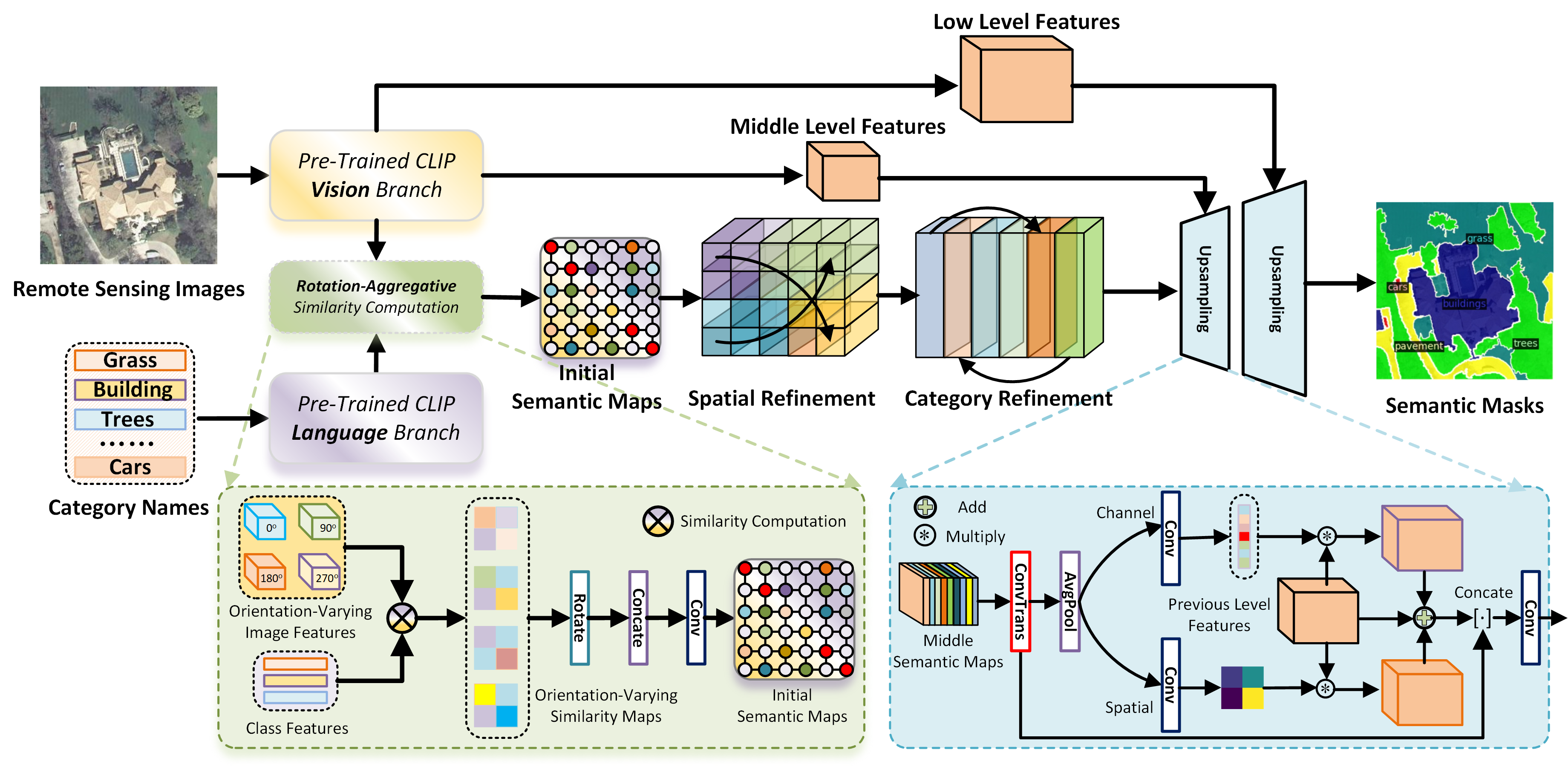}\\
  \caption{The overall framework of the proposed open-vocabulary remote sensing image semantic segmentation. Query images are initially rotated at multiple angles to generate orientation-specific image features using the vision branch of CLIP as the feature extractor. Simultaneously, category names are passed through the language branch to derive text embeddings, which serve as class features. By performing rotation-aggregative similarity computations between the orientation-specific image features and class features, the initial semantic maps are generated, capturing orientation-adaptive semantics. These maps are further refined spatially and categorically to enhance their precision. Additionally, to address scale variations, features from different levels are integrated during the upsampling process to progressively refine the semantic maps, leading to the final scale-aware semantic masks. }\label{fig2}
  \end{center}
\end{figure*}

\section{Proposed Method}


\subsection{Method Overview}
Consider an image \( I \) and a set of potential class categories \( C = \{T(n)\} \) where \( n = 1, \dots, N_C \), with \( T(n) \) representing the textual description of the \( n \)-th category and \( N_C \) being the total number of classes. For instance, \( T(n) \) could be defined as ``an image of \(Category\) \( Name \)''. The task of open-vocabulary semantic segmentation requires assigning a label to each pixel in the image \( I \) based on these categories. Unlike traditional semantic segmentation tasks, open-vocabulary segmentation faces the unique challenge of working with a flexible and dynamic set \( C \) provided as free-form text.

Following previous methods~\cite{liu2024open,xie2024sed,cho2024cat}, we employ the pre-trained CLIP model as the foundational vision-language framework to leverage its generalized knowledge. The overall structure of our proposed method is depicted in Figure~\ref{fig2}. Specifically, given query remote sensing images, they are first rotated through various orientations and then passed through the vision branch of the pre-trained CLIP model to extract orientation-specific image features. Concurrently, category names are input into the CLIP language branch to generate corresponding class features. Subsequently, in the rotation-aggregative similarity computation module, the network computes the similarities between the image and class features to produce orientation-specific similarity maps, effectively addressing the challenge of rapidly changing orientations. These similarity maps are subsequently aligned, concatenated, and fused to generate initial semantic maps. These maps undergo sequential spatial and category refinements to produce more accurate middle semantic maps. To further address significant scale variations, multi-level image features are incorporated to enrich the intermediate semantic maps with finer details that may have been overlooked due to scaling differences. During the upsampling process, these intermediate maps are used to activate preceding level features across both channel and spatial dimensions, enabling the network to better focus on relevant semantic regions. Ultimately, through successive upsampling stages, the network outputs the final, precise semantic masks.

\subsection{Rotation-Aggregative Similarity Computation}
Objects in remote sensing images tend to exist in varying orientations. Existing OVS methods often neglect this trait, and fail to tackle the varying orientations. Thus, we argue that the semantics from varying orientations should be gathered to ensure objects with varying orientations can be accurately parsed. Particularly, given the query remote sensing image $I_{q1}$, the image would be firstly rotated in diverse orientations, for instance, ($90^{\circ}$, $180^{\circ}$, and $270^{\circ}$), to acquire orientation-varying images $\left\{I_{q1}, I_{q2},..., I_{qN_A}\right\}$. These images are further propagated into the vision branch $P_V$ of Pre-trained CLIP to obtain the diverse image features from different angles $\left\{F_{q1},F_{q2},..., F_{qN_A}\right\} \in R^{HW \times d} $:

\begin{equation}
    F_{qi}=P_V(I_{qi}),i \in {1,2,...,N_A},
\end{equation}
Then, these image features  are  concatenated in a single dimension to obtain the orientation-vary image features $F_{qo} \in R^{HW \times N_A \times d}$:
\begin{equation}
   F_{qo}  = [F_{q1},F_{q2}, ...,F_{N_A}] 
\end{equation}

Meanwhile, the category names $\left\{C_{1}, C_{2}, ..., C_{N_C}\right\}$ would be inputted into the language branch $P_L$ of pre-trained CLIP to acquire the corresponding text embeddings as the class features $\left\{E_{1}, E_{2}, ..., E_{N_C}\right\} \in R^{1 \times d}  $:
\begin{equation}
    E_{j}=P_L(C_{j}),j \in {1,2,...,N_C},
\end{equation}

Subsequently, the orientation-varying semantics could be obtained through the cosine similarities between class features and orientation-varying image features:
\begin{equation}
{S_{qo}}[:,:,i] = \frac{{{F_{qo}}{E_j}}}{ \left\| {F_{qo}} \right\|  \left\| {E_j} \right\| },
\end{equation}
where ${S_{qi}} \in {R^{HW \times N_A \times N_C}}$ denotes the corresponding semantic maps. For better subsequent semantic understanding, these maps would be processed along the category dimension with a single convolutional layer:
\begin{equation}
    S^{'}_{qo}[:,:,i] = Conv(M_{qi}[:,:,i]) 
\end{equation}
where $S^{'}_{qo} \in R^{HW \times N_A \times N_C \times d_F} $. Then, the maps are correspondingly rotated along the angle dimension $N_A$ to the original orientation of $I_{q1}$ and further fused in the angle dimension to rotation-aggregately obtain the initial semantic maps $M_{q} \in R^{HW  \times N_C \times d_F} $:
\begin{equation}
    M_{q} = Conv([S^{'}_{qi}[:,0,:,:],...,Rotate(S^{'}_{qi}[:,N_A,:,:])])
\end{equation}
In this manner, the generated semantic maps adaptively aggregate semantic information across various orientations and successfully handle the rapidly changing orientations.

\subsection{Spatial and Category Refinement}
The acquired initial semantic maps $M_{q}$ are the coarse perceptions of query remote sensing images within the image-text semantic space of the pre-trained CLIP. To refine this coarse result, the semantic maps should be better analyzed respectively in both vision and text modalities. 

Firstly, to refine the holistic vision understanding of the semantic maps, the semantic maps should be sliced at the category dimension $N_c$, which helps the network concentrate on the vision level, and perform the spatial refinement at the pixel level.  We adopt the swin transformer~\cite{liu2021swin} as the refinement layer. More specifically, the refinement could be defined as follows:
\begin{equation}
    M_q^{'}[:,i,:] = L_T^{sp}(M_q[:,i,:]),i \in 1,2,...,N_C
\end{equation}
where $M_q[:,i,:] \in R^{(H \times W) \times d_F}$, and $L_T^{sp}$ denotes a pair of two consecutive Swin transformer layers for spatial refinement. Notably, $d_F$ is the channel dimensions for each token, and attention is performed on individual categories respectively.

Furthermore, the semantic maps would be refined at the text modality and the category refinement would help the network to precisely capture the relationships between multiple categories. Correspondingly, the semantic maps $M_q^{'}$ should be sliced at the spatial dimension $HW$ to eliminate the effect of vision modality. We also adopt the swin transformer as the refinement layer, and the process is defined as follows:
\begin{equation}
    M_q^{''}[i,:,:] = L_T^{ca}(M_q^{'}[i,:,:]),i \in 1,2,...,HW
\end{equation}
where $M_q^{'}[:,i,:] \in R^{N_C \times d_F}$. Different from the spatial refinement, there exist no spatial relations between categories, thus a linear transformer layer $L_T^{ca}$ without position embeddings is adopted for the category refinement. The overall refinement including both spatial and category refinements would repeat several times to acquire the middle semantic maps with higher accuracy.

\subsection{Scale-Aware Upsampling}
Although the rotation-aggregative similarity computation effectively enables the network to handle rapidly changing orientations, and the spatial and category refinement generates more precise semantic maps, significant scale variations remain unaddressed. To tackle this issue, it is crucial to incorporate multiscale image features from the feature extractor to enhance the model's scale adaptation ability. Specifically, since we use the pre-trained CLIP vision branch as the feature extractor, the image features from various layers of this branch are collected to provide vital scale information. The inclusion of these multiscale features ensures that the intermediate semantic maps capture previously neglected details, making the model more robust to scale variations.

Particularly, in the upsampling process, given the middle semantic maps $M_m\in R^{HW\times N_C \times d_F}$, and the previous level features $F_L \in R^{d_F \times H_L \times W_L}$, the middle semantic maps are firstly respectively average pooled respectively in spatial and channel levels, and go through the corresponding convolutional layers to obtain the spatial and channel activation vectors, which could be defined as:
\begin{equation}
{V_{sp}} = Conv(\mathop {Avgpool}\limits_{spatial} (Conv({M_m})))
\end{equation}
\begin{equation}
{V_{ch}} = Conv(\mathop {Avgpool}\limits_{channel} (Conv({M_m})))
\end{equation}
where $V_{sp} \in R^{HW\times1}$, and $V_{ch} \in R^{d_F\times1}$. Though the dot-product $\odot$ between these vectors and the previous level features $F_L$, the relevant semantic regions would be activated to obtain the spatial activated and channel activate features:
\begin{equation}
F_L^{sp} = {V_{sp}} \odot {F_L}
\end{equation}
\begin{equation}
F_L^{ch} = {V_{ch}} \odot {F_L}
\end{equation}
Through accumulating these activation features, and further concatenating with the middle semantic maps, the scale information is added into the semantic maps to obtain the scale-aware middle semantic maps $M^s_m$:
\begin{equation}
M_m^s = Conv([F_L^{sp} + F_L^{ch} + {F_L},{M_m}]),
\end{equation}

To prepare for the subsequent upsampling process, the intermediate semantic maps \( M^s_m \) are further processed through a single convolutional layer, acting as the connection layer. The upsampling process is repeated several times until the semantic maps are restored to their original scale. At each step, the previous image features are leveraged to provide scale information, ensuring that the network remains sensitive to variations in scale throughout the process. In this way, the final generated semantic masks \( M_F \) become both rotation-aggregative and scale-adaptive, enabling precise open-vocabulary segmentation in remote sensing imagery. The final generated semantic masks \( M_F \) are supervised by the ground truth mask \( M_{GT} \) using cross-entropy loss, defined as: 
\begin{equation}
\mathcal{L} = - \sum M_{GT} \log(M_F)
\end{equation}
This supervision encourages the model to produce accurate segmentation predictions by minimizing the difference between the predicted and actual semantic masks.

\begin{table*}[t!]
\caption{Categories for iSAID, DLRSD, Potsdam, and Vaihingen datasets}
	\centering
 	\renewcommand{\arraystretch}{1.5}
\resizebox{\linewidth}{!}{
\begin{tabular}{m{1.5cm}|m{12cm}}
\hline
\centering  Dataset &  \centering Category Names \arraybackslash  \\ \hline
\centering  iSAID & \centering ship, storage tank, baseball diamond, tennis court, basketball court, ground track field, bridge, large vehicle, small vehicle, helicopter, swimming pool, roundabout, soccer ball field, plane, harbor  \arraybackslash \\ \hline
\centering  DLRSD & \centering  airplane, bare soil, buildings, cars, chaparral, court, dock, field, grass, mobile home, pavement, sand, sea, ship, tanks, trees, water \arraybackslash \\ \hline
\centering  Potsdam & \centering impervious surfaces, Building, Low vegetation, Tree, Car, background \arraybackslash \\ \hline
\centering  Vaihingen &  \centering impervious surfaces, Building, Low vegetation, Tree, Car, background \arraybackslash \\ \hline
\end{tabular}
}
\label{tab1}
\end{table*}

\begin{table*}[t!]
\caption{Performance comparisons of different methods with different datasets as the training datasets. Val Data denotes the validation dataset of the chosen training dataset.}
	\centering
    \small
	\renewcommand{\arraystretch}{2.5}
	\renewcommand{\tabcolsep}{1.8mm}		
	\scalebox{0.9}{
\begin{tabular}{ccccccccccccccccc}
\hline
\multicolumn{2}{c|}{\multirow{2}{*}{Method}} &\multicolumn{3}{c|}{Val Data} & \multicolumn{3}{c|}{DLRSD} & \multicolumn{3}{c|}{iSAID}      & \multicolumn{3}{c|}{Potsdam}  & \multicolumn{3}{c}{Vaihingen} \\ \cline{3-17} 
\multicolumn{2}{c|}{}  & mIoU  & fwIoU  & \multicolumn{1}{c|}{mACC} & \multicolumn{1}{c}{mIoU}  & \multicolumn{1}{c}{fwIoU}  & \multicolumn{1}{c|}{mACC} &  \multicolumn{1}{c}{mIoU}  & \multicolumn{1}{c}{fwIoU}  & \multicolumn{1}{c|}{mACC} &  \multicolumn{1}{c}{mIoU}  & \multicolumn{1}{c}{fwIoU}  & \multicolumn{1}{c|}{mACC} & \multicolumn{1}{c}{mIoU}  & \multicolumn{1}{c}{fwIoU}  & \multicolumn{1}{c}{mACC} \\ \hline

\multicolumn{17}{c}{\textbf{DLRSD as training dataset}.}   \\ \hline

\multicolumn{2}{c|}{SCAN~\cite{liu2024open}}                   &  57.82  & 66.38  & \multicolumn{1}{c|}{75.01} &48.52 &55.01 &\multicolumn{1}{c|}{68.68} & 34.18 &44.62 &\multicolumn{1}{c|}{49.05} & 20.22 &28.37 &\multicolumn{1}{c|}{34.70} &5.38 &7.72 &22.54 \\
\multicolumn{2}{c|}{SAN~\cite{xu2023side}}                   & 83.28  & 88.32  & \multicolumn{1}{c|}{89.55}  & 85.73 & 86.24  &  \multicolumn{1}{c|}{91.03}  & 30.63  & 39.87 &  \multicolumn{1}{c|}{44.03} &  \textbf{30.30} &\textbf{38.04} & \multicolumn{1}{c|}{\textbf{44.98}} &31.92 &40.23 &  45.36 \\

\multicolumn{2}{c|}{SED~\cite{xie2024sed}}                    & 90.43 &92.03 &\multicolumn{1}{c|}{94.48}  & 85.13 &86.80 &\multicolumn{1}{c|}{91.36}  &21.54 &25.72 &\multicolumn{1}{c|}{36.28} &19.47 &21.29 &\multicolumn{1}{c|}{33.40}  &29.40 &36.77 &49.38\ \\
\multicolumn{2}{c|}{CAT-Seg~\cite{cho2024cat}}                  & 90.94 &93.07 &\multicolumn{1}{c|}{95.22}  &85.84 &86.79 &\multicolumn{1}{c|}{91.44} &23.56 &27.40 &\multicolumn{1}{c|}{38.48} &26.79 &31.28 &\multicolumn{1}{c|}{44.72} &32.32 &38.71 &49.65 \\

\multicolumn{2}{c|}{\textbf{Ours}}  & \textbf{91.11} &\textbf{93.22} &\multicolumn{1}{c|}{\textbf{95.45}} &\textbf{85.98 }&\textbf{86.94} &\multicolumn{1}{c|}{\textbf{91.52}} &\textbf{39.09} &\textbf{48.97} &\multicolumn{1}{c|}{\textbf{54.43}} &27.47 &26.81 &\multicolumn{1}{c|}{42.07}  &\textbf{33.71} &\textbf{38.57} &\textbf{50.01}   \\ \hline

\multicolumn{17}{c}{\textbf{iSAID as training dataset}.}    \\ \hline
\multicolumn{2}{c|}{SCAN~\cite{liu2024open}}                   & 64.59 &70.15 &\multicolumn{1}{c|}{76.73} &17.09 &19.50 &\multicolumn{1}{c|}{41.11} &60.37 &62.41 &\multicolumn{1}{c|}{74.43} &19.41 &25.91 &\multicolumn{1}{c|}{40.17} &9.88 &13.16 &26.40\\
\multicolumn{2}{c|}{SAN~\cite{xu2023side}}                   & 75.72 &82.72 &\multicolumn{1}{c|}{84.62} &20.28 &19.20 &\multicolumn{1}{c|}{\textbf{47.95}} &85.67 &88.02 &\multicolumn{1}{c|}{90.99}  &14.79 &17.69 &\multicolumn{1}{c|}{36.51} &16.38 &17.53 &35.38\\

\multicolumn{2}{c|}{SED~\cite{xie2024sed}}                    & 77.25 &84.70 &\multicolumn{1}{c|}{87.97} &\textbf{25.45} &24.09 &\multicolumn{1}{c|}{44.54} &92.92 &92.16 &\multicolumn{1}{c|}{95.49} & 15.27 &18.96 &\multicolumn{1}{c|}{28.57} &12.82 &9.61 &29.37\\
\multicolumn{2}{c|}{CAT-Seg~\cite{cho2024cat}}                  & 80.78 &86.64 &\multicolumn{1}{c|}{89.01}  &22.13 &23.93 &\multicolumn{1}{c|}{44.21} & 92.33 &92.51 &\multicolumn{1}{c|}{95.19} &14.68 &14.40 &\multicolumn{1}{c|}{37.47} &9.93 &10.87 &38.40\\

\multicolumn{2}{c|}{\textbf{Ours}}  &\textbf{87.87} &\textbf{89.07} &\multicolumn{1}{c|}{\textbf{92.30}} &20.67 &\textbf{26.18} &\multicolumn{1}{c|}{45.44}  &\textbf{93.29} &\textbf{92.52} &\multicolumn{1}{c|}{\textbf{95.77}} &\textbf{19.94} &\textbf{26.29} &\multicolumn{1}{c|}{\textbf{49.12}}  &\textbf{20.82} &\textbf{26.61} & \textbf{50.60}\\ \hline
\end{tabular}}
	\label{tab2}	
\end{table*}

\begin{table*}[t!]
\caption{Mean Performance comparisons of different methods. The best results are shown in bolded.}
	\centering
    \small
	\renewcommand{\arraystretch}{2.1}
	\renewcommand{\tabcolsep}{6.0mm}		
	\scalebox{1.0}{
\begin{tabular}{cccccccc}
\hline
\multicolumn{2}{c|}{\multirow{2}{*}{Method}} &\multicolumn{3}{c|}{DLRSD as training dataset} & \multicolumn{3}{c}{iSAID as training dataset} \\ \cline{3-8} 
\multicolumn{2}{c|}{}  & mIoU  & fwIoU  & \multicolumn{1}{c|}{mACC} & \multicolumn{1}{c}{mIoU}  & \multicolumn{1}{c}{fwIoU}  & \multicolumn{1}{c}{mACC} \\ \hline

\multicolumn{2}{c|}{SCAN~\cite{liu2024open}}                   &  33.22  & 40.42  & \multicolumn{1}{c|}{49.99} &34.26 &38.22 & 51.77  \\
\multicolumn{2}{c|}{SAN~\cite{xu2023side}}                   & 52.37  & 58.54  & \multicolumn{1}{c|}{62.99}  & 42.57 & 45.03  &  59.09 \\

\multicolumn{2}{c|}{SED~\cite{xie2024sed}}                    & 49.19 &52.52 &\multicolumn{1}{c|}{60.98}  & 44.74 &45.90 &57.19 \\
\multicolumn{2}{c|}{CAT-Seg~\cite{cho2024cat}}                  & 51.89 &55.45 &\multicolumn{1}{c|}{63.90}  &43.97 &45.67 &60.85 \\
\multicolumn{2}{c|}{\textbf{Ours}}                  & \textbf{55.47} &\textbf{58.90} &\multicolumn{1}{c|}{\textbf{66.70}}  &\textbf{48.52} &\textbf{52.13} &\textbf{66.65}  \\

\hline
\end{tabular}}
	\label{tab3}	
\end{table*}

\section{Experiments}

To thoroughly evaluate the effectiveness of the proposed method, we conduct an extensive set of experiments on the newly introduced open-vocabulary remote sensing semantic segmentation benchmark. The experimental section is organized as follows. First, we provide a detailed description of the datasets used and the evaluation metrics adopted. Next, we outline the key implementation details of our approach. Following this, a comprehensive analysis of the performance comparison between our method and state-of-the-art OVS approaches is presented, both in qualitative and quantitative terms. Finally, we perform a series of ablation studies to assess the contribution of each component of our method.

\subsection{Dataset and Evaluation Metric}

To comprehensively evaluate the proposed method, we expand the experimental setup beyond the commonly used Potsdam and Vaihingen semantic segmentation datasets by including the processed DLRSD~\cite{chaudhuri2017multilabel} and iSAID~\cite{yao2021scale} datasets, thereby establishing a more robust open-vocabulary semantic segmentation (OVS) benchmark for remote sensing imagery. The processed DLRSD dataset consists of 7002 images across 17 categories, while the processed iSAID dataset includes 24439 images across 15 categories. Potsdam and Vaihingen, which share the same six categories, contain 20102 and 2254 images, respectively. Detailed category names for these datasets are provided in Table~\ref{tab1}. 

In our experimental setup, DLRSD and iSAID are used as the training datasets, while evaluations are conducted across all datasets to assess the OVS performance. Specifically, 5601 images from the DLRSD dataset and 1401 images from iSAID are used for training, with the remaining 1401 and 6363 images used for validation, respectively. Following standard evaluation protocols from prior OVS methods~\cite{cho2024cat,xie2024sed}, we utilize Mean Intersection over Union (mIoU), Frequency Weighted Intersection over Union (fwIoU), and Mean Accuracy (mACC) as evaluation metrics, providing a comprehensive reflection of the model's performance.

The mIoU is computed as the average IoU across all classes:
\begin{equation}
\text{IoU}_i = \frac{TP_i}{TP_i + FP_i + FN_i}, \quad \text{mIoU} = \frac{1}{N} \sum_{i=1}^{N} \text{IoU}_i,
\end{equation}
where \(TP_i\), \(FP_i\), and \(FN_i\) represent true positives, false positives, and false negatives for class \(i\), and \(N\) is the total number of classes.

The fwIoU metric accounts for the frequency of each class in the dataset:
\begin{equation}
\text{fwIoU} = \frac{1}{\sum_{i=1}^{N} n_i} \sum_{i=1}^{N} n_i \cdot \frac{TP_i}{TP_i + FP_i + FN_i},
\end{equation}
where \(n_i\) is the number of pixels in class \(i\).

Finally, mACC measures the per-class accuracy, defined as the ratio of correctly classified pixels to the total number of pixels for each class:
\begin{equation}
\text{ACC}_i = \frac{TP_i}{TP_i + FN_i}, \quad \text{mACC} = \frac{1}{N} \sum_{i=1}^{N} \text{ACC}_i.
\end{equation}

\begin{table*}[t!]
\caption{ Performance comparisons of diverse classes on the Vaihingen dataset. The best results are shown in bolded.}
    \centering
    \normalsize
    \renewcommand{\arraystretch}{1.6}
    \renewcommand{\tabcolsep}{2.0mm}        
    \scalebox{0.95}{
\begin{tabular}{cl|>{\centering\arraybackslash}m{2cm}|>{\centering\arraybackslash}m{2cm}|>{\centering\arraybackslash}m{2cm}|>{\centering\arraybackslash}m{2cm}|>{\centering\arraybackslash}m{2cm}|>{\centering\arraybackslash}m{2cm}|>{\centering\arraybackslash}m{2cm}}
\hline
\multicolumn{2}{c|}{Methods} & \multicolumn{1}{c|}{impe. sur.} & \multicolumn{1}{c|}{Building} & \multicolumn{1}{c|}{Low vegetation} & \multicolumn{1}{c|}{Tree} & \multicolumn{1}{c|}{Car} & \multicolumn{1}{c|}{background} & \multicolumn{1}{c}{mean IoU} \\ \hline
\multicolumn{9}{c}{DLRSD as the training dataset} \\   \hline 
\multicolumn{2}{c|}{SCAN~\cite{liu2024open}}                   & 13.69 &13.93 &0.86 &0.23 &2.51 &1.06 &5.38  \\
\multicolumn{2}{c|}{SAN~\cite{xu2023side}}                   &\textbf{42.59} &58.50 &14.75 &41.50 &34.03 &0.15  &31.92  \\

\multicolumn{2}{c|}{SED~\cite{xie2024sed}}                    & 41.29 &50.75 &\textbf{15.36} &36.40 &29.43 &3.15  &29.40  \\
\multicolumn{2}{c|}{CAT-Seg~\cite{cho2024cat}}               &37.54 &65.22 &9.20 &38.05 &36.71 &7.19 & 32.32 \\
\multicolumn{2}{c|}{\textbf{Ours}}  &37.27 &\textbf{66.07} &2.50 &\textbf{42.35} &\textbf{44.77} &\textbf{9.29}  &\textbf{33.71}  \\ 

 \hline 
\multicolumn{9}{c}{iSAID as the training dataset}    \\ \hline

\multicolumn{2}{c|}{SCAN~\cite{liu2024open}}                   & 14.54 &15.92 &6.81 &14.93 &5.87 &1.24 &9.88  \\
\multicolumn{2}{c|}{SAN~\cite{xu2023side}}                   & 21.65 &0.62 &\textbf{21.34} &28.22 &25.17 &1.29  &16.88  \\

\multicolumn{2}{c|}{SED~\cite{xie2024sed}}                    & \textbf{28.79} &0.30 &4.48 &0.12 &\textbf{41.26} &1.95    &12.82  \\
\multicolumn{2}{c|}{CAT-Seg~\cite{cho2024cat}}                  & 26.28 &1.40 &2.83 &10.17 &15.99 &0.93 &9.93  \\
\multicolumn{2}{c|}{\textbf{Ours}}  &21.79 &\textbf{36.69} & 2.80 &\textbf{43.85} &16.23 &\textbf{3.56}  &\textbf{20.82}  \\  \hline
\end{tabular}}
\label{tab4}    
\end{table*}

\subsection{Implement Details}
We utilize the pre-trained CLIP ViTB/16~\cite{dosovitskiy2020image} as the foundational vision-language feature extractor, leveraging its ability to provide robust visual and textual embeddings. The proposed method is implemented using PyTorch and Detectron2 frameworks, ensuring flexibility and efficiency in large-scale experimentation. For optimization, we adopt the AdamW optimizer~\cite{loshchilov2017decoupled} with an initial learning rate of 2e-4 to balance convergence speed and stability, and a weight decay of 10e-4 to prevent overfitting. The training is performed with a batch size of 4, utilizing an Nvidia A100 GPU with 80GB of memory to manage the computational demands, especially given the high-resolution remote sensing imagery. All training images are resized to $3 \times 384 \times 384$ to maintain consistency across datasets and to align with the input requirements of the CLIP model. The models are trained for a total of 100,000 iterations, ensuring ample convergence time for learning the intricate features required for open-vocabulary semantic segmentation in remote sensing imagery. After training on the processed DLRSD/iSAID datasets, the resulting models are thoroughly evaluated on both their respective validation sets and the complete benchmark of four remote sensing datasets (Potsdam, Vaihingen, DLRSD, and iSAID) to provide a comprehensive assessment of their generalization capabilities.

\begin{figure*}[t!]
  \begin{center}
  \includegraphics[width=1.0\linewidth]{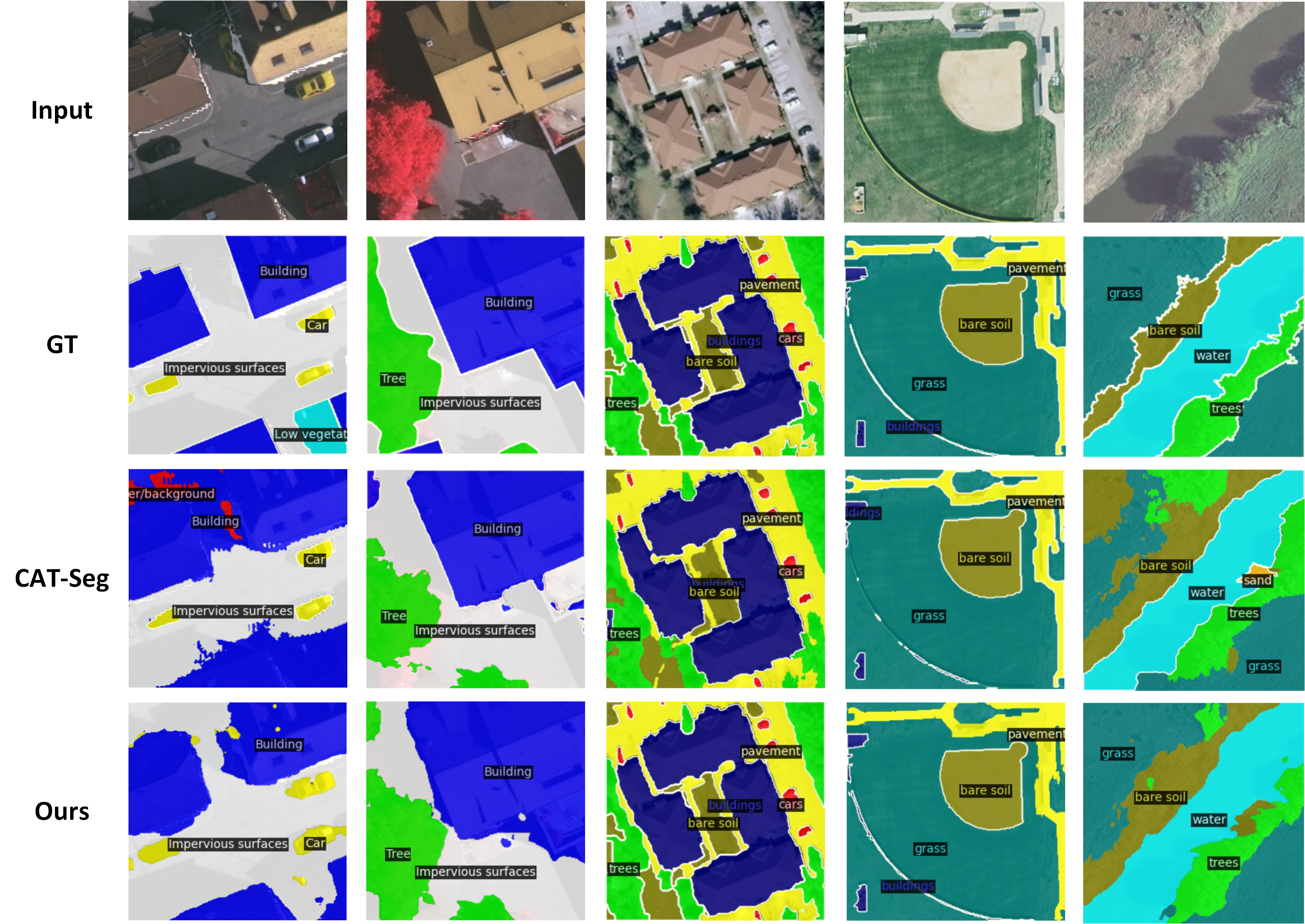}\\
  \caption{Qualitative results of the proposed method. From top to bottom: Input images, the ground truth of query images, predictions of CAT-Seg, and predictions of our methods. }\label{fig3}
  \end{center}
\end{figure*}

\subsection{Performance Analysis}
The quantitative performance comparisons between our proposed method and other advanced OVS algorithms are detailed in Table~\ref{tab2}, with the highest results highlighted in bold. When trained on DLRSD, our method outperforms others on most metrics. Notably, compared with the state-of-the-art CAT-Seg method, our approach shows substantial improvements on the iSAID dataset, with an increase of 15.53$\%$ in mIoU, 11.57$\%$ in fwIoU, and 15.95$\%$ in mACC. Across the validation datasets of DLRSD, Potsdam, and Vaihingen, our method also demonstrates superior performance with consistent and significant gains. However, it falls short of the top performance on the Potsdam dataset compared to the SAN method, though it still ranks second in mIoU.  This underperformance may be attributed to a potential dataset domain gap between DLRSD dataset and Potsdam dataset, which could hamper the segmentation performance on Potsdam. Addressing this dataset domain gap remains a challenge for future work.

In contrast, when trained on the larger iSAID dataset, our method achieves the best results across nearly all metrics. Most notably, the Vaihingen dataset sees dramatic performance gains, with mIoU and fwIoU both surpassing 20$\%$, while the mACC of our method reaches 50.60$\%$. These results underscore the generalization capability of our method when given a diverse and extensive dataset for training. Nevertheless, on the DLRSD dataset, our approach fails to surpass all metrics, specifically mIoU and mACC, suggesting the persistent gap between DLRSD and iSAID data distributions, which reflects the inherent variability in remote sensing data sources.

To further understand the performance differences, we also provide average performance comparisons in Table~\ref{tab2}. Notably, our method outperforms others in all metrics, regardless of whether DLRSD or iSAID is used for training. Interestingly, with a relatively small DLRSD as the training dataset, our method still achieves the best average results though fails in Potsdam, indicating that DLRSD images may offer more representative features for diverse categories. In particular, the highest metric improvements are seen in mACC, where a 2.80$\%$ boost was achieved, bringing the overall mACC to 66.70$\%$. These results reaffirm the effectiveness of our approach, emphasizing its capability to leverage the unique traits of remote sensing imagery. This significant performance improvement further highlights the necessity of tailoring OVS algorithms to the specific challenges of earth observation, setting a new benchmark for OVS in remote sensing.

To delve deeper into the performance, a detailed breakdown of class-wise performance comparisons on the Vaihingen dataset is presented in Table~\ref{tab4}. Overall, our proposed method demonstrates superior performance, achieving 33.71$\%$ mIoU when trained on DLRSD and 20.82$\%$ mIoU with iSAID as the training dataset. Specifically, using DLRSD for training, our method shows the best results on the building class with a remarkable 66.07$\%$ mIoU, while the most significant performance gain is observed on the car class, showing an improvement of approximately 8$\%$ mIoU. When trained on the iSAID dataset, our method performs exceptionally well on the tree class, with a leading 43.85$\%$ mIoU, and achieves the most substantial performance boost on the building class, showing a striking near 21$\%$ mIoU improvement.  One intriguing observation is that although all OVS methods struggle with segmenting pixels annotated as background, our method still manages to show improvements in this area. This suggests that the incorporation of remote sensing-specific traits meaningfully contributes to better semantic understanding in earth observation tasks. These improvements underscore the ability of our approach to adapt to the unique challenges of remote sensing imagery and successfully push the boundaries of open-vocabulary segmentation in this domain.

To further evaluate the effectiveness of the proposed method, we visualize the segmentation results, as shown in Figure~\ref{fig3}. Our method clearly achieves superior performance by accurately segmenting regions and improving the quality of predictions. Notably, the rotation-aggregative similarity computation module significantly enhances the model's ability to parse remote sensing objects, reducing misclassification. For example, in the first and second columns, our method correctly classifies more pixels as impervious surfaces and demonstrates greater accuracy in identifying building areas compared to the CAT-Seg model. 

Moreover, the scale-aware upsampling process proves to be a key factor in enhancing semantic detail parsing. In the third and fourth columns, where other methods wrongly classify detailed pixels as pavement or grass, our method successfully corrects these misclassifications, identifying them as bare soil and buildings, respectively. These qualitative results highlight the importance of incorporating unique traits of remote sensing imagery, enabling our model to achieve more precise semantic segmentation.

\begin{figure}[t!]
  \begin{center}
  \includegraphics[width=1.0\linewidth]{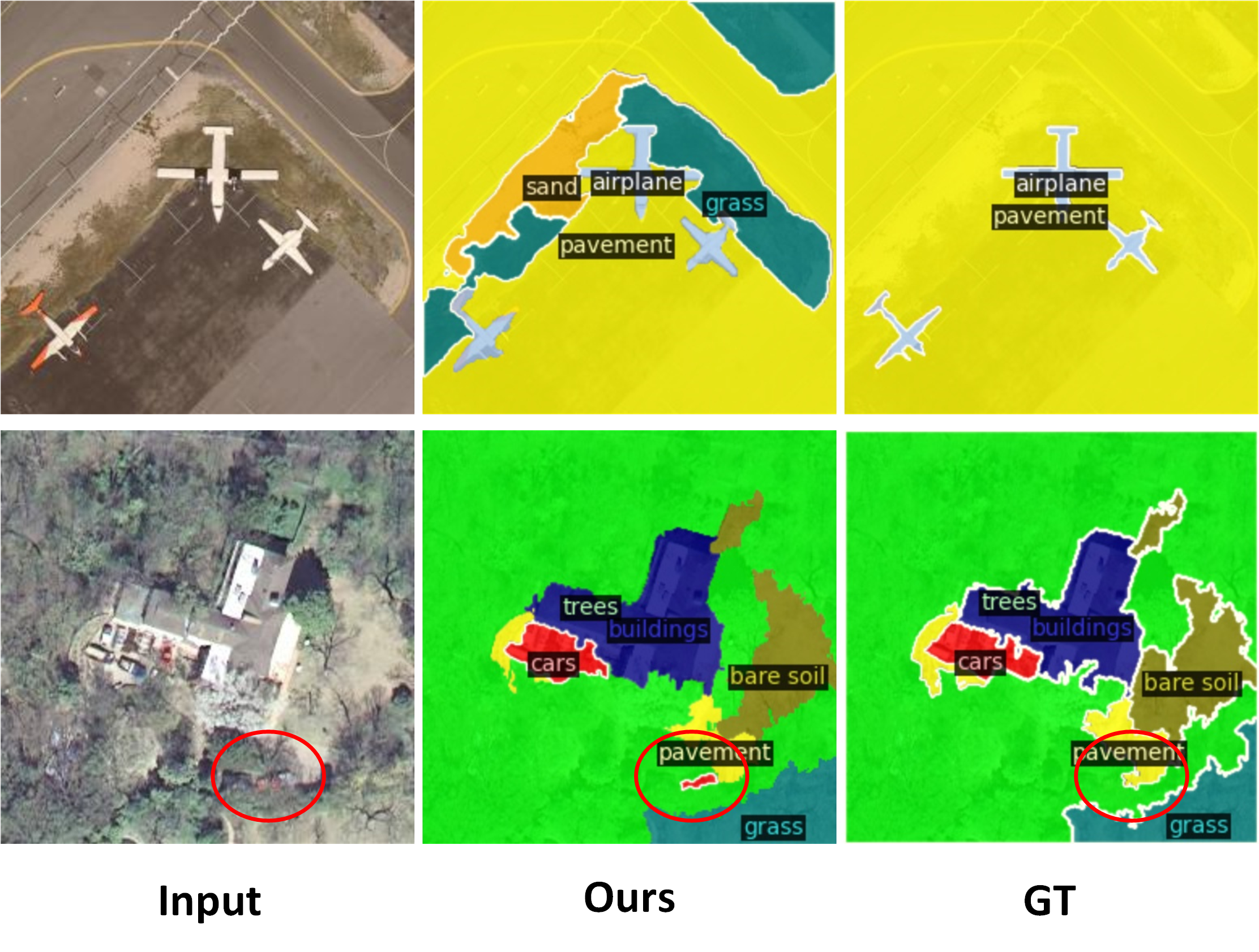}\\
  \caption{Some intriguing visualization results of our proposed method. }\label{fig4}
  \end{center}
\end{figure}

In addition, when analyzing the segmentation results further, we observed some noteworthy outcomes, as illustrated in Figure~\ref{fig4}. Interestingly, in certain cases, the ground truth contains annotation errors that our method manages to correct. For instance, in the first row, several pixels are incorrectly labeled as pavement in the ground truth, but our method accurately classifies some of them as grass and sand, a distinction validated by the input image. A similar pattern can be seen in the second row, where our model correctly segments pixels annotated as pavement in the ground truth as cars, as confirmed by the input image. 

These findings demonstrate that our method possesses a true open-vocabulary capability, going beyond predefined category lists. The model can recognize and label categories that were not explicitly annotated in the training data, underscoring the value of incorporating open-vocabulary segmentation in remote sensing tasks.

\begin{table}[t]
	\centering
    \small
	\renewcommand{\arraystretch}{2.1}
	\renewcommand{\tabcolsep}{3.7mm}
	\caption{Ablation study of the rotation-aggregative similarity computation and scale-aware upsampling. RSC illustrates the rotation-aggregative similarity computation, and SAU denotes the scale-aware upsampling. }
	\scalebox{1.0}{
		\begin{tabular}{cc|cccc}
			\hline
			\multicolumn{1}{c}{\multirow{1}{*}{RSC}}&  \multicolumn{1}{c|}{\multirow{1}{*}{SAU}} &\multicolumn{1}{c}{mIoU}  &\multicolumn{1}{c}{fwIoU}  &\multicolumn{1}{c}{mACC} &\multicolumn{1}{c}{Mean} \\
			\hline
			\multicolumn{1}{c}{} &    \multicolumn{1}{c|}{}  &80.78 &86.65 &89.02 &85.48  \\ 
			\multicolumn{1}{c}{\checkmark} & \multicolumn{1}{c|}{} &87.06 &89.03 & 91.22 & 89.10  \\
   			\multicolumn{1}{c}{} & \multicolumn{1}{c|}{\checkmark} &85.81 &88.47 &90.64 &88.31  \\
			\multicolumn{1}{c}{\checkmark} &  \multicolumn{1}{c|}{\checkmark} &\textbf{87.87} &\textbf{89.07} &\textbf{92.30} &\textbf{89.75}  \\  \hline		
	\end{tabular}}
	\label{tab5}
\end{table}

\begin{table}[t]
	\centering
     \small
	\renewcommand{\arraystretch}{2.1}
	\renewcommand{\tabcolsep}{3.8mm}
	\caption{The ablation study of diverse orientations. }
	\scalebox{0.9}{
\begin{tabular}{cl|cccc}
\hline
\multicolumn{2}{c|}{Orientations}      & mIoU        & fwIoU        & mACC        & Mean           \\ \hline
\multicolumn{2}{c|}{$[0^{\circ}]$}             &80.78 &86.65 &89.02 &85.48      \\
\multicolumn{2}{c|}{$[0^{\circ}$,$90^{\circ}]$}         & 85.79         & 87.79          & 90.93          & 88.17          \\
\multicolumn{2}{c|}{[$0^{\circ}$,$90^{\circ}$,$180^{\circ}]$}     & 86.18          & 88.13          & 91.09          & 88.47          \\
\multicolumn{2}{c|}{[$0^{\circ}$,$90^{\circ}$,$180^{\circ}$,$270^{\circ}]$} &\textbf{87.06} &\textbf{89.03} &\textbf{91.22} &\textbf{89.10} \\ 
\multicolumn{2}{c|}{[$0^{\circ}$,$45^{\circ}$,...,$270^{\circ}$,$315^{\circ}$]} &  86.86	&88.63	&90.91	&88.80         \\ 
\hline
\end{tabular}}
	\label{tab6}
\end{table}

\subsection{Ablation Study}

To evaluate the effectiveness of our proposed OVS framework for remote sensing imagery, a series of experiments are performed to analyze the effect of the key components. All ablation experiments are conducted with the iSAID as the training dataset, and the corresponding validation dataset as the evaluation dataset. 

First, we analyze the contributions of the rotation-aggregative similarity computation and the scale-aware upsampling process individually, with results summarized in Table~\ref{tab5}. Notably, the network's ability to handle varying orientations significantly improves segmentation performance, boosting the mIoU from 80.78$\%$ to 87.06$\%$. Addressing scale variations also leads to substantial gains, resulting in a performance increase of approximately 5$\%$. By jointly tackling these unique remote sensing traits, our proposed method achieves a new state-of-the-art performance. A deeper analysis reveals that handling orientation variations provides greater performance benefits compared to scale variations, which indicates that both orientations and scales are critical factors contributing to the final segmentation performance, yet the parsing results benefit more from the orientations.

In our framework, we rotate input images at three angles ($90^\circ$, $180^\circ$, $270^\circ$) along with the original image to account for orientation-varying objects. To study the impact of aggregating multiple orientations in the rotation-aggregative similarity computation module, the experimental results are shown in Table~\ref{tab6}. As expected, increasing the number of orientations improves segmentation performance incrementally. A significant performance gain is observed when introducing the first new orientation, further reinforcing the importance of handling rapidly changing orientations. The network achieves peak performance when aggregating $0^\circ$, $90^\circ$,  $180^\circ$, and $270^\circ$. Beyond this, adding more orientations slightly hampers performance, likely due to redundant information and the increased parameter count, which could interfere with the extraction of rotation-adaptive features. Therefore, the combination of these four primary orientations appears to strike the optimal balance, functioning effectively as the four axes of an orientation system for generating rotation-aggregative features.

Moreover, we conduct experiments to analyze the number of layers from which features are introduced to address the rapidly changing scale variations. Sequentially incorporating features from earlier layers improves segmentation performance. Specifically, the best results are achieved when features from two previous layers are included. However, incorporating features from additional shallower layers slightly degrades performance. This suggests that excessively shallow features may lack sufficient scale information and contain overly coarse semantics, which are not beneficial for the upsampling process in remote sensing imagery. Consequently, introducing features from two layers strikes the optimal balance between providing scale-aware details and maintaining semantic coherence for accurate segmentation.

Overall, the ablation studies clearly demonstrate that both the rotation-aggregative similarity computation and the scale-aware upsampling process are essential to the success of the proposed method. By addressing the unique orientation and scale challenges inherent to remote sensing imagery, the method significantly enhances segmentation performance, achieving state-of-the-art results. Additionally, the careful selection of orientation aggregation and feature layers proves crucial in optimizing the network's ability to adapt to these challenges, ensuring more accurate and robust semantic segmentation in open-vocabulary settings. These insights not only validate the importance of the proposed components but also provide a foundation for future improvements in remote sensing OVS.

\begin{table}[t]
	\centering
     \small
	\renewcommand{\arraystretch}{2.1}
	\renewcommand{\tabcolsep}{3.8mm}
	\caption{Ablation study of the layer number of the scale-aware upsampling process.}
	\scalebox{1.0}{
\begin{tabular}{c|cccc}
\hline
Layer Number & mIOU   & fwIoU        & mACC &Mean         \\ \hline
0   &80.78 &86.65 &89.02 &85.48    \\
1    &84.66  & 87.89  & 90.17 &87.57\\
2  &\textbf{85.81} &\textbf{88.47} &\textbf{90.64} &\textbf{88.31}  \\
3 &85.62 &88.11 &90.23 &87.99  \\ \hline
\end{tabular}}
\label{tab7}
\end{table}


\section{Conclusion}
In this paper, we introduced the first open-vocabulary semantic segmentation (OVS) benchmark for remote sensing imagery and developed a novel open-sourced OVS framework specifically tailored for earth vision. To address the unique challenges of remote sensing imagery, such as rapidly changing orientations and significant scale variations, we proposed a rotation-aggregative similarity computation module that generates orientation-adaptive similarity maps. These maps are subsequently refined at spatial and categorical levels to enhance segmentation accuracy. Additionally, we integrated multi-scale image features into the upsampling process to produce scale-aware semantic masks, enabling the framework to handle significant scale variations. Extensive experiments on four public remote sensing datasets under open-vocabulary settings demonstrated that our method achieves state-of-the-art performance, affirming its effectiveness in earth vision tasks. Looking forward, this work lays the foundation for future advancements in OVS for remote sensing, opening avenues for exploring more robust handling of other remote sensing vision perception challenges.




%





\ifCLASSOPTIONcaptionsoff
  \newpage
\fi





\bibliographystyle{IEEEtran}
\bibliography{IEEEabrv,Bibliography}
%

\begin{IEEEbiography}[{\includegraphics[width=1in,height=1.25in,clip,keepaspectratio]{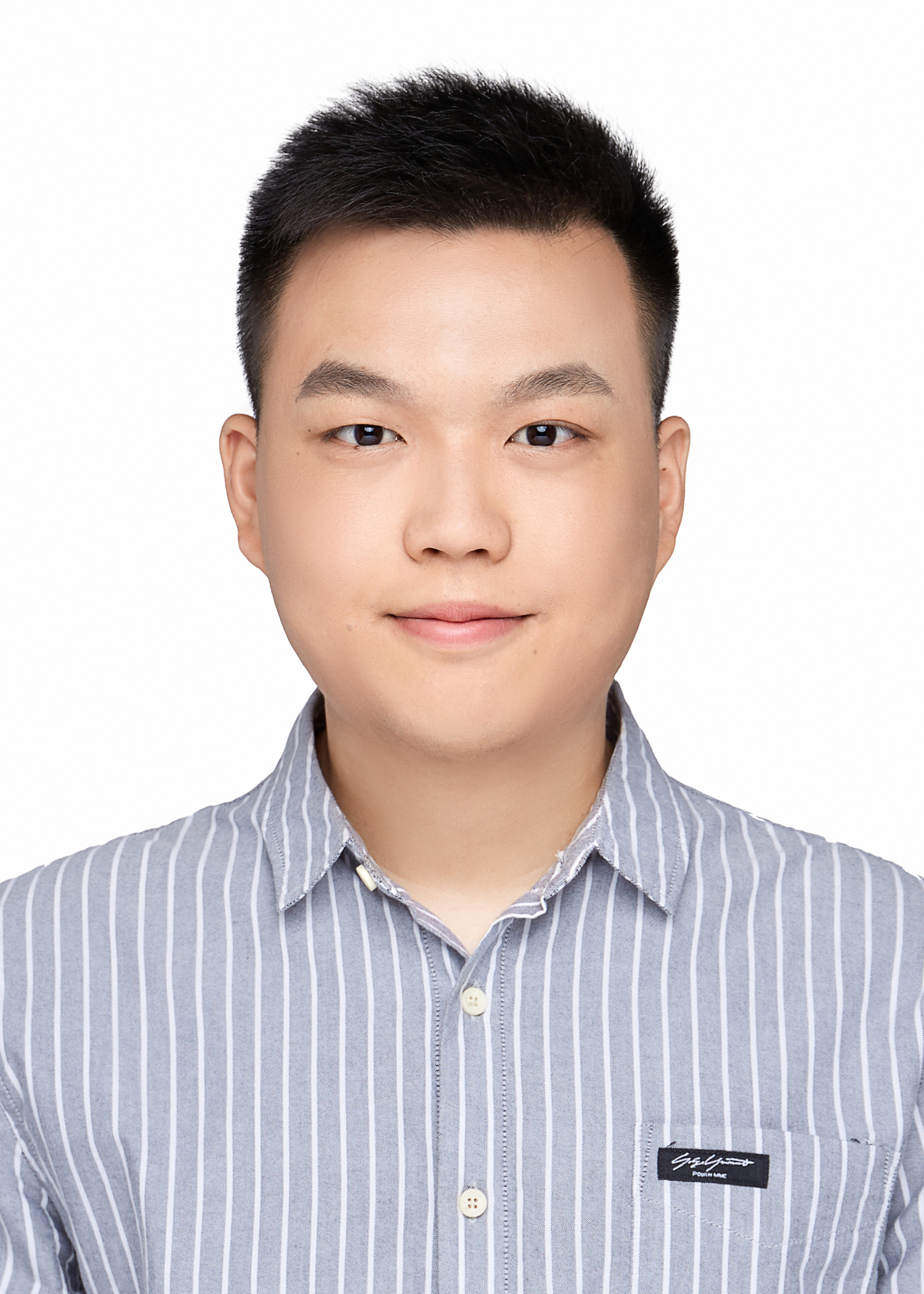}}]{Qinglong Cao} received his B.E. and M.S. degrees from Northwestern Polytechnical University, Xi’an, China, in 2019 and 2022, respectively. He is currently working toward the Ph.D. degree at Shanghai Jiao Tong University. His research interests include computer vision and remote sensing image processing, especially on few-shot learning and semantic segmentation.
\end{IEEEbiography}

\begin{IEEEbiography}[{\includegraphics[width=1in,height=1.25in,clip,keepaspectratio]{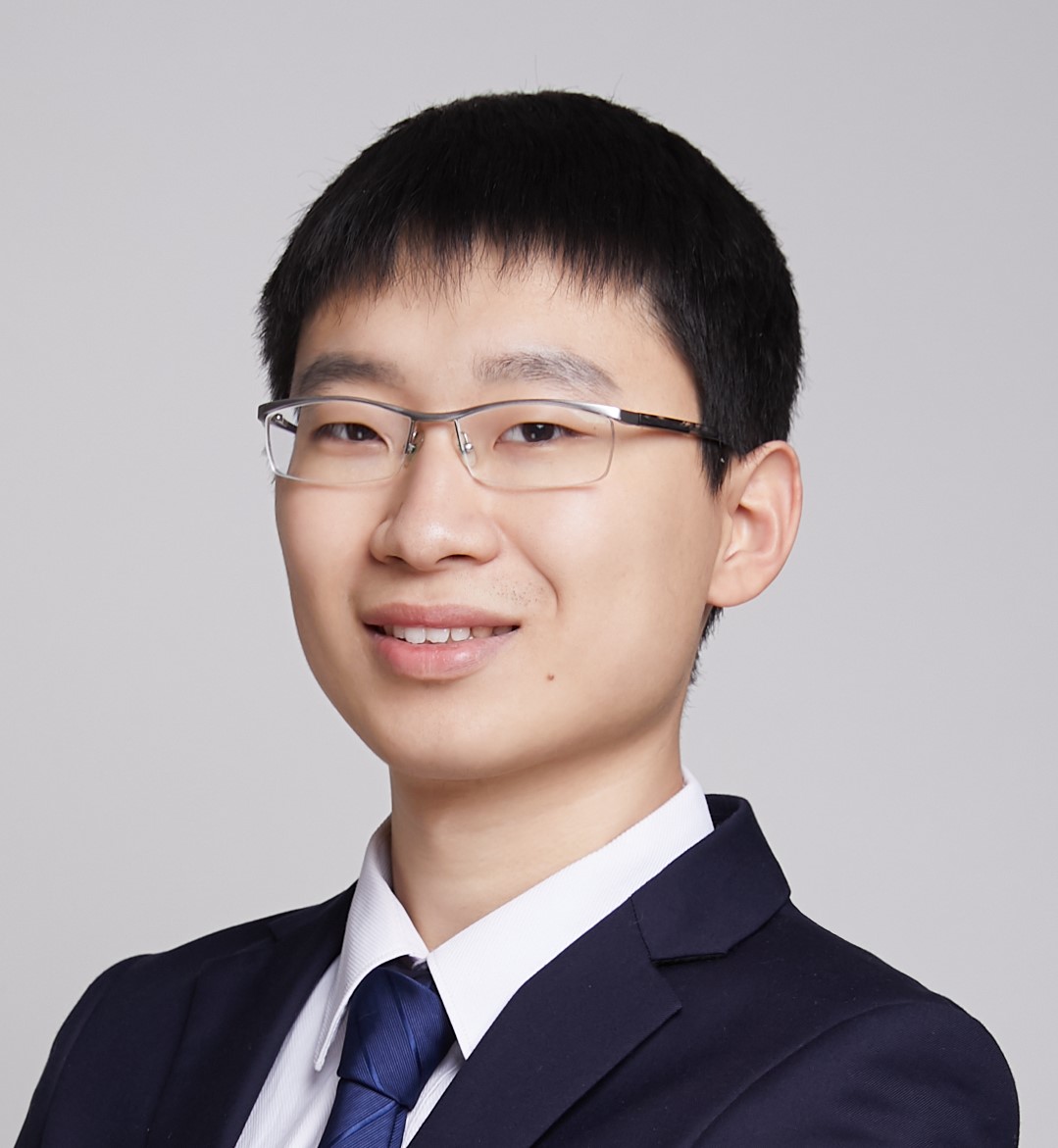}}]{Yuntian Chen} is an assistant professor at Eastern Institute of Technology, Ningbo.  He received the B.S. degree from Tsinghua University, Beijing, China, in 2015, the dual B.S. degree from Peking University, Beijing, China, in 2015, and the Ph.D. degree with merit from  Peking University, Beijing, China, in 2020.  His research field includes scientific machine learning and intelligent energy systems. He is interested in the integration of domain knowledge and data-driven models.
\end{IEEEbiography}

\begin{IEEEbiography}[{\includegraphics[width=1in,height=1.25in,clip,keepaspectratio]{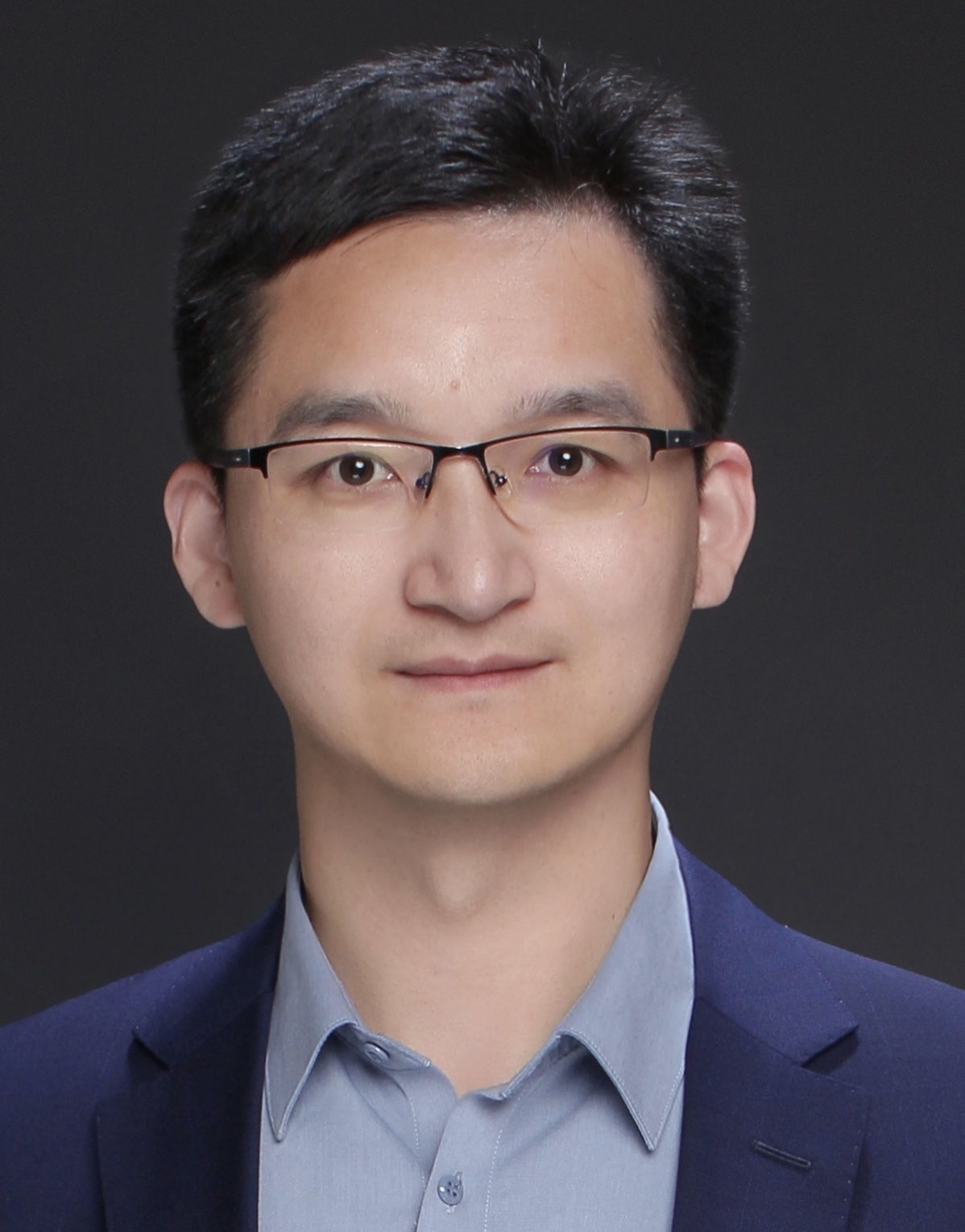}}]{Chao Ma}
(Member, IEEE) is an associate professor at Shanghai Jiao Tong University, Shanghai, 200240, P. R. China. He received the Ph.D. degree from Shanghai Jiao Tong University in 2016. His research interests include computer vision and machine learning. He was a research associate with School of Computer Science at The University of Adelaide from 2016 to 2018. He was sponsored by China Scholarship Council as a visiting Ph.D. student at University of California at Merced from the fall of 2013 to the fall of 2015.
\end{IEEEbiography}

\begin{IEEEbiography}[{\includegraphics[width=1in,height=1.25in,clip,keepaspectratio]{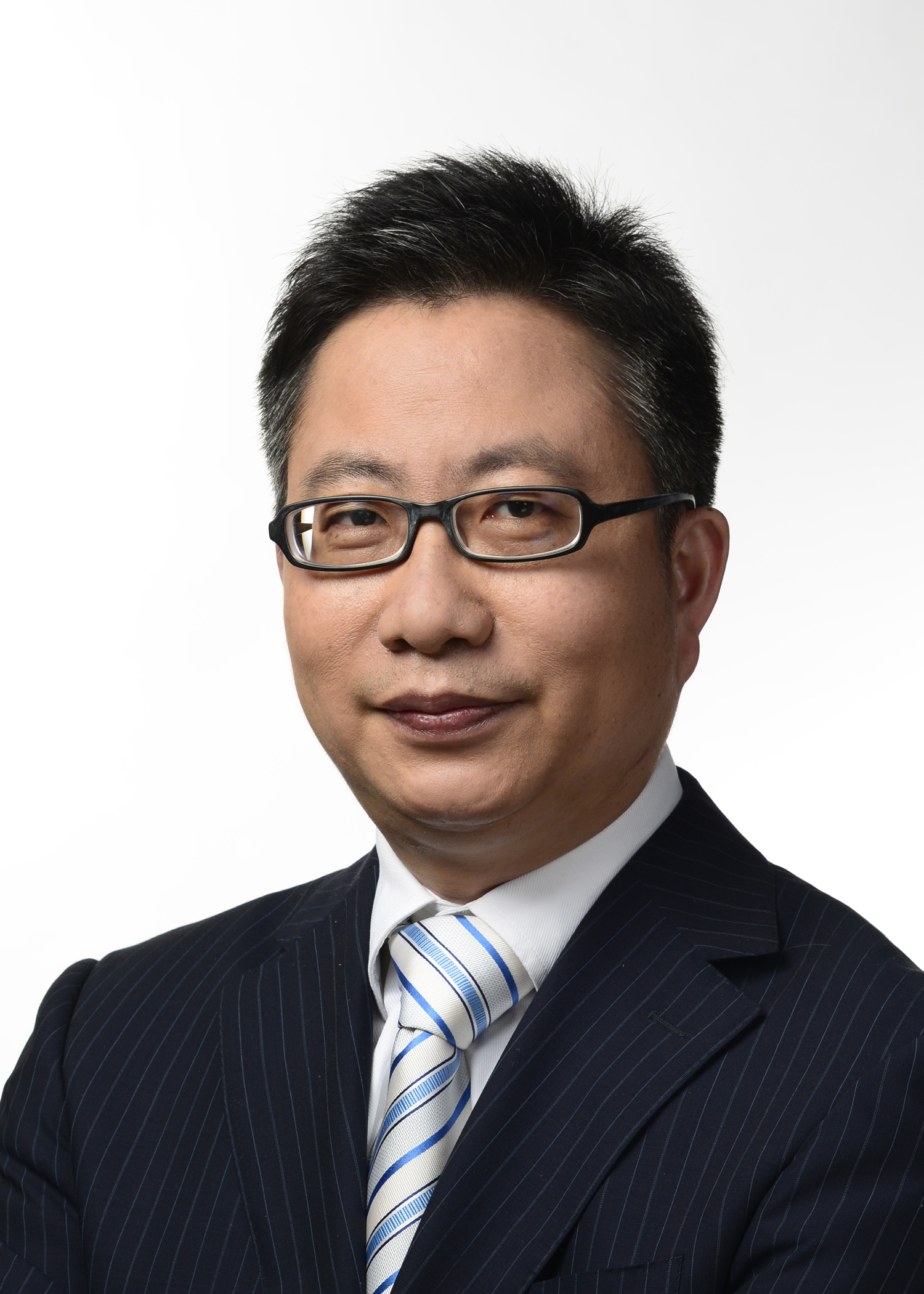}}]{Xiaokang Yang}
(Fellow, IEEE) received the B.S. degree from Xiamen University, Xiamen, China, in 1994, the M.S. degree from the Chinese Academy of Sciences, Shanghai, China, in 1997, and the Ph.D. degree from Shanghai Jiao Tong University, Shanghai, in 2000. From September 2000 to March 2002, he worked as a Research Fellow with the Centre for Signal Processing, Nanyang Technological University, Singapore. From April 2002 to October 2004, he was a Research Scientist at the Institute for Infocomm Research (I2R), Singapore. From August 2007 to July 2008, he visited the Institute for Computer Science, University of Freiburg, Germany, as an Alexander von Humboldt Research Fellow. He is currently a Distinguished Professor at the School of Electronic Information and Electrical Engineering, Shanghai Jiao Tong University. He has published over 200 refereed articles and has filed 60 patents. His current research interests include image processing and communication, computer vision, and machine learning. He received the 2018 Best Paper Award of the IEEE Transactions on Multimedia. He is an Associate Editor of the IEEE Transactions on Multimedia and a Senior Associate Editor of the IEEE Signal Processing Letters.
\end{IEEEbiography}





\vfill


\end{document}